\documentclass[11pt]{article}

\usepackage[preprint]{acl}

\usepackage{times}
\usepackage{latexsym}
\usepackage[T1]{fontenc}
\usepackage[utf8]{inputenc}
\usepackage{microtype}
\usepackage{graphicx}
\usepackage{amsmath}
\usepackage{amssymb}
\usepackage{booktabs}
\usepackage{multirow}
\usepackage{tabularx}

\title{SAGE: A Hierarchical Framework for Evaluating Interpretive Literary Quality in Narratives}

\author{Tianyu Wang \\
  Mercy University \\
  Math \& Computer Science Department \\
  Dobbs Ferry, NY, USA \\
  \texttt{twang4@mercy.edu} \\\And
  Nianjun Zhou \\
  IBM T.J.\ Watson Research Center \\
  Yorktown Heights, NY, USA \\
  \texttt{jzhou@us.ibm.com} \\}

\begin{document}
\maketitle

\begin{abstract}
Assessing the literary quality of narratives requires evaluating interpretive dimensions (cultural representation, emotional depth, and philosophical engagement) that existing NLG metrics cannot measure. We introduce SAGE, a six-layer evaluation framework that separates rule-based assessment of observable textual properties from LLM-based evaluation of interpretive qualities drawn from cultural theory, affect theory, and existentialist philosophy. Each interpretive layer is assessed through multi-round iterative LLM evaluation with independent cross-validation, achieving measurement-grade reliability (98.8\% convergence, $>$94\% inter-rater agreement) stable across evaluator models. Validated on 600 evaluations across 100 short stories, our central finding is a systematic capability boundary: emotional-psychological representation approaches human levels, while cultural critique and philosophical depth exhibit approximately double the gap. LLM-generated narratives score below even commercial genre fiction on all three layers. We interpret this as a boundary between \emph{pattern-reproducible} literary capacities learnable from training corpora and \emph{stance-requiring} ones demanding cultural positioning and philosophical engagement that pattern matching alone cannot provide.
\end{abstract}

\section{Introduction}
\label{sec:introduction}

Evaluating the literary quality of narratives requires measuring interpretive dimensions that existing NLG metrics cannot reach: cultural representation, emotional-psychological depth, and philosophical engagement. Standard metrics (BLEU, ROUGE, BERTScore; \citealt{papineni2002bleu,lin2004rouge,zhang2019bertscore}) measure surface similarity and correlate poorly with human judgments of narrative quality \citep{novikova2017we,reiter2018structured}. Narrative-specific frameworks extend evaluation to structural properties such as coherence and character consistency \citep{chhun2022human,yi2025score}, but these interpretive dimensions remain unaddressed. As large language models generate increasingly fluent narratives, the absence of reliable evaluation for these dimensions creates a systematic gap: generative systems receive no signal for improvement precisely where the deficits are largest.

We propose SAGE (Systematic Assessment of Generative Excellence) to address this blind spot. SAGE is a hierarchical evaluation framework that decomposes literary quality into six analytically distinct layers, separating rule-based assessment of observable textual properties (L1--L3) from LLM-based evaluation of interpretive dimensions (L4--L6). Each interpretive dimension is grounded in established theoretical frameworks from literary and cultural studies, and assessment proceeds through a dual-track architecture combining five-round iterative self-reflection with independent cross-validation.

We validate SAGE on 100 short stories across three interpretive layers in two evaluation modes. Applying SAGE reveals a systematic capability boundary: emotional-psychological representation is substantially closer to human levels than cultural representation and philosophical depth, with the latter two exhibiting approximately double the gap. We interpret this as a boundary between \emph{pattern-reproducible} literary capacities (learnable from surface regularities in training corpora) and \emph{stance-requiring} ones that demand genuine cultural positioning and philosophical engagement. LLM-generated narratives score below even commercial genre fiction on all three layers, suggesting the deficit reflects something more fundamental than insufficient literary training data.

These findings motivate four research questions:

\noindent\textbf{RQ1 (Reliability):} Can the multi-round LLM evaluation architecture produce stable, reproducible assessments of interpretive dimensions?

\noindent\textbf{RQ2 (Discriminative Capacity):} Does SAGE differentiate meaningfully among canonical literature, genre fiction, and LLM-generated narratives?

\noindent\textbf{RQ3 (Robustness):} Are evaluation outcomes stable across different informational conditions (text-based vs.\ reputation-based assessment)?

\noindent\textbf{RQ4 (Dimensional Independence):} Do the proposed dimensions capture non-redundant aspects of literary quality?

\paragraph{Contributions.}
This work makes two contributions. First, we demonstrate that theory-driven multi-round LLM evaluation can achieve measurement-grade reliability ($>$94\% inter-rater agreement, 98.8\% convergence) on interpretive literary dimensions, establishing LLM-as-judge as viable for tasks that previously resisted automated assessment. Second, we identify a systematic capability boundary: emotional representation is pattern-reproducible and approaches human competence, while cultural critique and philosophical depth are stance-requiring and exhibit approximately double the gap---a distinction invisible to existing NLG evaluation methods and now, for the first time, reliably measurable. Code, prompts, dataset, and results are publicly available.\footnote{\url{https://anonymous.4open.science/r/sageFramework-journal-EDD6/}}

\section{Related Work}
\label{sec:related}

Evaluating narrative quality is a long-standing challenge. Standard NLG metrics (BLEU, ROUGE, BERTScore; \citealt{papineni2002bleu,lin2004rouge,zhang2019bertscore}) measure surface similarity to reference texts and correlate poorly with human narrative quality judgments \citep{reiter2018structured,novikova2017we}. Recognizing this, narrative-specific frameworks have extended evaluation to structural dimensions: HANNA \citep{chhun2022human} provides human-annotated story evaluations across six criteria; NarraBench \citep{hamilton2025narrabench} surveys 78 benchmarks for narrative understanding; SCORE \citep{yi2025score} tracks character consistency and emotional coherence; and computational work has characterized discourse coherence \citep{barzilay2008modeling} and emotional trajectories \citep{reagan2016emotional}. Computational literary scholarship has further applied machine learning to trace genre evolution and literary prestige at scale \citep{underwood2019distant}, and positioned these methods within narratological theory \citep{piper2021narrative}. Together, these approaches establish that narrative evaluation requires decomposition beyond surface metrics, yet they remain focused on observable, structural properties. The interpretive dimensions that distinguish canonical literature from competent prose (how a text positions itself within cultural and ideological terrain, how it renders inner life, whether it sustains genuine philosophical inquiry) are not addressed by any existing framework.

Relevant NLP subfields illuminate each of these gaps individually, yet do not bridge them. Bias detection \citep{blodgett2020language} examines harmful representations but not the sophistication of cultural engagement as a literary quality; sentiment analysis \citep{poria2019emotion} classifies broad affective categories but not the emotional granularity and interiority that characterize sophisticated fiction; topic modeling \citep{blei2003latent} identifies recurring themes but not the depth of philosophical engagement. These tools address adjacent problems; none was designed to evaluate interpretive literary quality as such.

LLM-based evaluation has emerged as a scalable alternative to human annotation \citep{zheng2023judging,li2024llms}. Studies show strong LLMs achieve over 80\% agreement with human preferences, though systematic biases (position, verbosity, self-preference) require mitigation. Fine-grained frameworks such as FActScore \citep{min2023factscore} demonstrate that decomposing evaluation into explicit dimensions substantially improves precision. However, existing LLM-as-judge work targets conversational or instructional quality; creative and literary evaluation remains theoretically underspecified. Recent process-oriented analysis of LLM narrative generation suggests that models systematically prioritize stylistic surface over character, event, and thematic depth \citep{jung2025style}---a generation bias that mirrors the evaluation blind spot we identify. As LLM-generated narratives approach human-level fluency \citep{wu2025longeval}, the absence of reliable evaluation for these deeper dimensions becomes the critical bottleneck for NLG progress.

\section{Methodology}
\label{sec:methodology}

\subsection{Framework Overview}

\begin{figure*}[t]
\centering
\includegraphics[width=\textwidth]{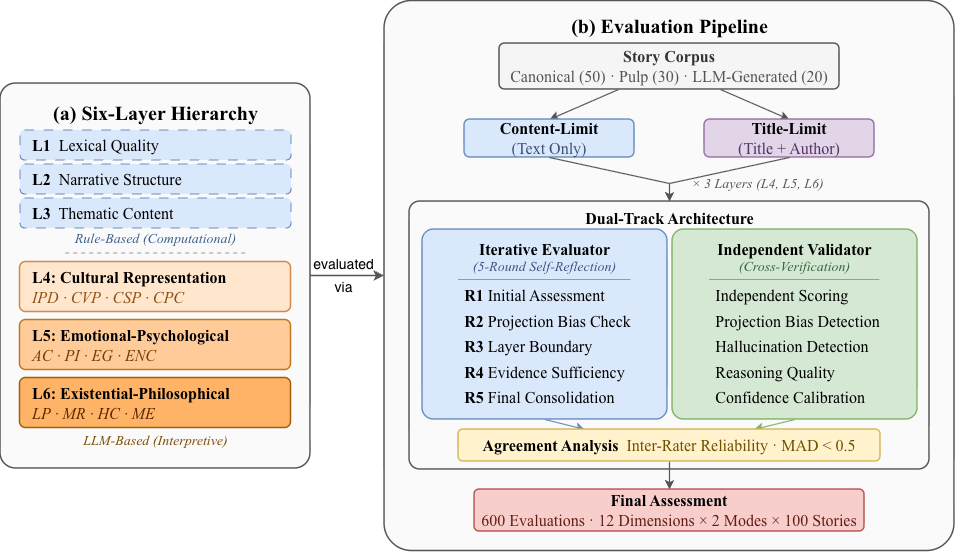}
\caption{SAGE architecture. Layers 1--3 assess textual properties through rule-based metrics; Layers 4--6 evaluate interpretive dimensions through a dual-track LLM pipeline (iterative evaluator + independent validator) in two modes.}
\label{fig:framework}
\end{figure*}

SAGE decomposes literary quality into six analytical layers organized in a progression from observable textual properties (L1--L3, rule-based) to increasingly interpretive literary qualities (L4--L6, LLM-based). The term \emph{hierarchical} here denotes this progression in analytical depth, not dependency between layers: L1--L3 and L4--L6 are evaluated independently, and no layer's scores are computed from another's. This reflects the hypothesis that formal features and interpretive qualities constitute largely independent dimensions requiring different analytical approaches. Figure~\ref{fig:framework} illustrates the architecture; Table~\ref{tab:framework_layers} summarizes the six layers. Our experimental validation focuses on L4--L6, which require reasoning and cultural knowledge available only through modern LLMs.

The three interpretive layers correspond to the three core questions that literary criticism has long identified as defining literary significance: how a text positions itself within social and cultural power structures (L4), how it renders the inner life of characters (L5), and how it engages with fundamental questions of human existence (L6). These are precisely the dimensions absent from existing NLG evaluation frameworks, and the dimensions where, as our results show, LLM-generated narratives fall furthest short.

\begin{table}[t]
\centering
\caption{Six-layer hierarchical evaluation framework.}
\label{tab:framework_layers}
\small
\resizebox{\columnwidth}{!}{%
\begin{tabular}{clp{4.2cm}}
\toprule
\textbf{L} & \textbf{Focus} & \textbf{Key Dimensions} \\
\midrule
\multicolumn{3}{l}{\textit{Rule-Based (L1--L3)}} \\
1 & Lexical & Vocabulary richness, syntactic complexity \\
2 & Narrative & Entity coherence, emotional arcs \\
3 & Thematic & Topic diversity, semantic networks \\
\midrule
\multicolumn{3}{l}{\textit{LLM-Based (L4--L6)}} \\
4 & Cultural & IPD, CVP, CSP, CPC \\
5 & Emotional & AC, PI, EG, ENC \\
6 & Existential & LP, MR, HC, ME \\
\bottomrule
\end{tabular}%
}
\end{table}

\subsection{LLM-Based Evaluation Layers (L4--L6)}

\subsubsection{Layer 4: Cultural Representation}

L4 evaluates cultural engagement and power structures, grounded in postcolonial and cultural theory \citep{said1978orientalism,spivak1988subaltern,geertz1973interpretation} and field theory \citep{bourdieu2019distinction}. Four dimensions operationalize this layer: \textbf{Intersectional Power Dynamics (IPD)}, the complexity of power relations across social axes; \textbf{Cultural Voice \& Perspective (CVP)}, how the narrative voice is positioned relative to represented cultures; \textbf{Cultural Specificity (CSP)}, the density of cultural particularity in the text; and \textbf{Cultural Pattern Complexity (CPC)}, engagement with multiple cultural logics simultaneously.

\subsubsection{Layer 5: Emotional-Psychological Representation}

L5 evaluates emotional and psychological representation, grounded in affect theory \citep{sedgwick1995shame,berlant2020cruel}, emotion granularity research \citep{barrett2017emotions}, and narrative psychology \citep{cohn1978transparent,wood2008fiction}. Four dimensions operationalize this layer: \textbf{Affective Complexity (AC)}, the multiplicity and contradiction of emotional states; \textbf{Psychological Interiority (PI)}, the depth of access to characters' inner worlds; \textbf{Emotional Granularity (EG)}, the precision and differentiation of emotional vocabulary; and \textbf{Emotional-Narrative Coherence (ENC)}, whether emotions arise organically from the narrative situation.

\subsubsection{Layer 6: Existential-Philosophical Representation}

L6 evaluates philosophical depth and existential engagement, grounded in existentialist philosophy \citep{heidegger2010being,sartre2022being,camus1969myth}, moral philosophy \citep{levinas1979totality,macintyre2013after}, and hermeneutics \citep{ricoeur1980narrative,gadamer2013truth}. Four dimensions operationalize this layer: \textbf{Life Philosophy (LP)}, engagement with fundamental questions of human existence; \textbf{Moral Reflection (MR)}, the depth of ethical inquiry and moral complexity; \textbf{Human Condition (HC)}, the exploration of mortality, suffering, and vulnerability; and \textbf{Meaning Exploration (ME)}, how narratives construct, question, or subvert frameworks of meaning.

\subsubsection{Construct Validity: Dimension Non-Redundancy}

The twelve dimensions are designed to be non-redundant: canonical works occupy divergent positions on theoretically motivated axes, establishing that dimensions dissociate in practice. Table~\ref{tab:exemplars} illustrates key orthogonalities across all three layers. At L5, Hemingway's ``Hills Like White Elephants'' achieves high AC (contradictory emotions embedded in subtext) while PI is near zero (no access to either character's inner states), the inverse pattern from Woolf's \emph{Mrs.\ Dalloway}. At L4, Conrad's \emph{Heart of Darkness} and Achebe's \emph{Things Fall Apart} share high IPD but occupy opposite CVP positions: outsider gaze versus insider cultural authority over the same colonial encounter. These dissociations confirm that the dimensions capture genuinely independent aspects of literary quality; full exemplar matrices are in Appendix~\ref{sec:appendix_exemplars}.

\begin{table}[t]
\centering
\caption{Selected exemplars demonstrating dimension non-redundancy across layers. H\,=\,High, M\,=\,Medium, L\,=\,Low. Dissociating pairs highlighted in bold.}
\label{tab:exemplars}
\small
\setlength{\tabcolsep}{4pt}
\resizebox{\columnwidth}{!}{%
\begin{tabular}{p{2.9cm}lll}
\toprule
\textbf{Work} & \textbf{Dims.} & \textbf{Values} & \textbf{Layer} \\
\midrule
``Hills Like White Eleph.'' & AC / PI & \textbf{H / L} & L5 \\
\emph{Mrs.\ Dalloway} & AC / PI & H / H & L5 \\
\emph{Heart of Darkness} & IPD / CVP & \textbf{H / L} & L4 \\
\emph{Things Fall Apart} & IPD / CVP & H / H & L4 \\
\emph{The Stranger} & LP / MR & \textbf{H / L} & L6 \\
\emph{Crime and Punishment} & LP / MR & H / H & L6 \\
\bottomrule
\end{tabular}%
}
\end{table}

\subsection{Dual-Mode Evaluation}

We implement two evaluation modes that vary the information provided to the model. \textbf{Content-limit mode} provides only the story text without title, author, or metadata, requiring the LLM to assess literary qualities from observable textual evidence alone. \textbf{Title-limit mode} provides only the work's title and author without text, requiring synthesis of scholarly consensus from the model's training corpus. Comparing these modes reveals whether LLM evaluations rely on textual analysis, memorized critical opinions, or both.

\subsection{Dual-Track Evaluation Architecture}

To ensure reliable assessments, we employ two evaluators with complementary roles.

\paragraph{Iterative Evaluator.} Assessment proceeds across five structured rounds. \textbf{Round 1} extracts relevant content and produces initial scores (1.0--5.0) with confidence ratings and textual evidence. \textbf{Round 2} applies layer-specific bias mitigation: L4 performs a hallucination check (fabricated cultural-historical claims); L5 checks projection bias (imposing Western emotional norms); L6 checks Western framework imposition (overlooking Buddhist, Confucian, or Daoist philosophical depth). \textbf{Round 3} verifies layer boundary compliance: L5 must not evaluate cultural power structures (L4) or philosophical claims (L6). \textbf{Round 4} re-examines evidence sufficiency, adjusting confidence (not scores) where evidence is sparse. \textbf{Round 5} consolidates final scores; convergence criterion: $|\Delta\text{score}|<0.3$ from Round~4 to Round~5.

\paragraph{Independent Validator.} A separate evaluator assesses the same texts without access to the iterative evaluator's outputs, providing (1) independent dimension scores, (2) projection bias and hallucination detection, (3) agreement analysis with iterative evaluator scores, and (4) an overall trust assessment. Agreement between evaluators provides inter-rater reliability evidence; systematic disagreements reveal where LLM evaluation proves less reliable.

All rounds produce structured JSON outputs. Complete prompt templates are provided in Appendix~\ref{sec:appendix_prompts}.

\section{Experiments}
\label{sec:experiments}

\subsection{Dataset}

We validate the framework through systematic evaluation of 100 short stories across three quality categories. To control for length effects, all stories fall within the 2,000--8,000 word range characteristic of the short fiction form.

\paragraph{Canonical Literature (n=50).} Short stories consistently recognized by academic institutions and literary scholarship, spanning authors including Chekhov, Kafka, Joyce, Faulkner, Borges, Lu Xun, Garc\'ia M\'arquez, Hemingway, Poe, and Woolf. Publication dates span 1835--1990, encompassing Romanticism, Realism, Modernism, and Postmodernism across Russian, European, North American, Latin American, and Japanese literary traditions.

\paragraph{Pulp Fiction (n=30).} Short stories published between 1880 and 1950 in commercial genre magazines (\textit{All-Story}, \textit{Weird Tales}, \textit{Black Mask}), including adventure, horror, western, detective, and early science fiction. This corpus provides a controlled comparison category distinguished from canonical literature by critical reception and commercial publication context, exhibiting technical narrative competence while typically employing genre conventions and formulaic structures.

\paragraph{LLM-Generated Stories (n=20).} Stories selected from the lars76/story-evaluation-llm dataset \citep{huggingface-story-eval}, containing narratives generated by Mistral, Gemma, and Llama variants. The corpus includes quality stratification: 10 high-quality stories (human evaluation scores 3.58--3.78 on a 5-point scale), 5 average-quality (score 3.40), and 5 low-quality (scores 2.56--2.58), enabling testing of whether SAGE tracks deliberate quality variation within AI-generated content.

\begin{table}[t]
\centering
\caption{Evaluation corpus and matrix.}
\label{tab:corpus}
\small
\resizebox{\columnwidth}{!}{%
\begin{tabular}{lccl}
\toprule
\textbf{Category} & \textbf{Stories} & \textbf{Evals.} & \textbf{Quality} \\
\midrule
Canonical & 50 & 300 & Academically recognized \\
Pulp Fiction & 30 & 180 & Commercial genre \\
LLM-Generated & 20 & 120 & Multi-model, stratified \\
\midrule
\textbf{Total} & \textbf{100} & \textbf{600} & 3 layers $\times$ 2 modes \\
\bottomrule
\end{tabular}%
}
\end{table}

\subsection{Evaluation Procedure}

Each story undergoes evaluation across L4, L5, and L6 in both modes (content-limit and title-limit), each performed by both the iterative evaluator (5 rounds) and independent validator (1 round). This yields 600 complete evaluations (Table~\ref{tab:corpus}).

All LLM-based evaluations employ GPT-5-mini (OpenAI) with reasoning effort set to ``medium'' and structured JSON outputs. Prompts incorporate explicit theoretical frameworks for each layer (Bourdieu/Said/Geertz for L4; Sedgwick/Barrett/Wood for L5; Heidegger/Levinas/Ricoeur for L6), establish evidence sovereignty (all scores must cite textual evidence), enforce strict layer boundaries, and apply layer-specific bias mitigation as described in Section~\ref{sec:methodology}. Prompt templates are provided in Appendix~\ref{sec:appendix_prompts}.

\subsection{Statistical Analysis}

RQ1 reliability is assessed via convergence rates ($|\Delta|<0.3$ from Round~4 to Round~5) and inter-rater MAD. RQ2 discriminative validity uses independent-samples $t$-tests and one-way ANOVA with Bonferroni correction; effect sizes reported as Cohen's $d$. RQ3 robustness uses paired $t$-tests comparing content-limit and title-limit scores. RQ4 dimensional independence uses Pearson and Spearman correlations across 200 story-mode observations.

\section{Results}
\label{sec:results}

\subsection{Reliability and Robustness (RQ1, RQ3)}

The multi-round architecture performs as designed across all 600 evaluations (100\% success rate). Scores stabilize by Rounds~3--4, with 98.8\% convergence ($|\Delta|<0.3$ from R4 to R5; trajectories in Appendix~\ref{sec:appendix_convergence}), and independent validation yields $>$94\% inter-rater agreement (mean MAD $<$0.3). The dual-mode comparison tells a similar story: content-limit and title-limit scores differ by at most 0.05 points overall ($p>0.05$), and stratified analysis confirms this holds within each category (max 0.09 points; all $p>0.05$)---including LLM-generated stories, where title-limit identifiers carry no literary signal, ruling out reputation retrieval as a confound. Table~\ref{tab:reliability} summarizes these metrics.

\begin{table}[t]
\centering
\caption{System reliability metrics across 600 evaluations.}
\label{tab:reliability}
\small
\resizebox{\columnwidth}{!}{%
\begin{tabular}{lcc}
\toprule
\textbf{Metric} & \textbf{Value} & \textbf{Threshold} \\
\midrule
Evaluation success rate & 100.0\% & --- \\
Convergence rate (R4$\to$R5) & 98.8\% & $|\Delta|<0.3$ \\
Inter-rater agreement & $>$94\% & MAD $<$ 0.5 \\
Mode invariance (max $|\Delta|$) & 0.05 & $p>0.05$ \\
\bottomrule
\end{tabular}%
}
\end{table}

\subsection{Genre Discrimination (RQ2)}

\begin{figure}[t]
\centering
\includegraphics[width=\columnwidth]{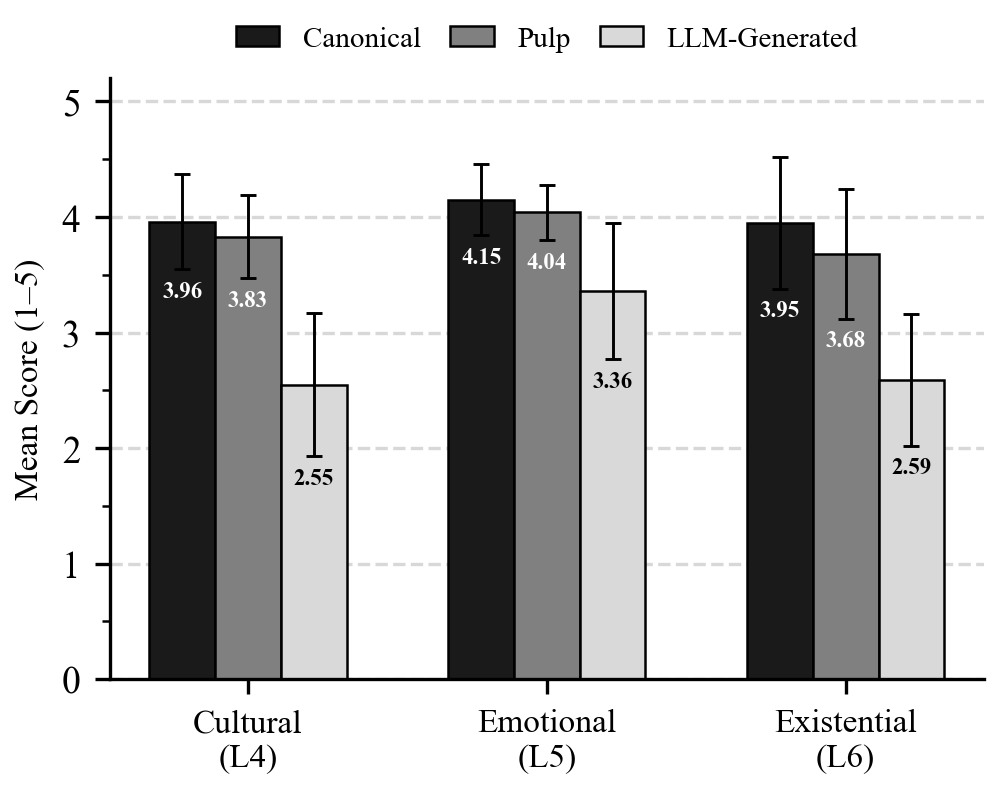}
\caption{Genre comparison across three analytical layers. Error bars indicate standard deviations. All canonical vs.\ LLM-generated differences significant at $p<0.001$.}
\label{fig:genre}
\end{figure}

SAGE identifies a clear and statistically significant quality hierarchy across all three layers: Canonical (4.02) $>$ Pulp (3.85) $>$ LLM (2.83), with all canonical-vs.-LLM pairwise comparisons significant at $p<0.001$ (Figure~\ref{fig:genre}, Table~\ref{tab:genre}). The overall gap of +1.19 points (29.6\% relative) is unambiguous, but the more informative finding is in the layer-by-layer breakdown. Effect sizes are not uniform: L4 ($d$=2.68) and L6 ($d$=2.40) are roughly 60\% larger than L5 ($d$=1.68), indicating qualitatively different capability profiles rather than uniform underperformance.

\begin{table}[t]
\centering
\caption{Genre scores by layer (mean $\pm$ SD) with Cohen's $d$ and $p$-values for Canonical vs.\ LLM pairwise comparison. Canon.\ vs.\ Pulp: all layers $p>0.05$ (n.s.); Pulp vs.\ LLM: all layers $p<0.001$.}
\label{tab:genre}
\small
\resizebox{\columnwidth}{!}{%
\begin{tabular}{lccccc}
\toprule
\textbf{Layer} & \textbf{Canon.} & \textbf{Pulp} & \textbf{LLM} & \textbf{$d$} & \textbf{$p$ (C vs L)} \\
\midrule
L4 Cultural     & 3.96{\small$\pm$.41} & 3.83{\small$\pm$.36} & 2.55{\small$\pm$.62} & 2.68 & $<$0.001 \\
L5 Emotional    & 4.15{\small$\pm$.31} & 4.04{\small$\pm$.24} & 3.36{\small$\pm$.59} & 1.68 & $<$0.001 \\
L6 Existential  & 3.95{\small$\pm$.57} & 3.68{\small$\pm$.56} & 2.59{\small$\pm$.57} & 2.40 & $<$0.001 \\
\midrule
\textbf{Average} & \textbf{4.02} & \textbf{3.85} & \textbf{2.83} & --- & --- \\
\bottomrule
\end{tabular}%
}
\end{table}

The pulp fiction results sharpen this picture. Commercial authors writing under genre constraints nonetheless achieve emotional scores approaching canonical literature (4.04 vs.\ 4.15, a gap of just 2.7\%), yet LLM-generated stories fall \emph{below} pulp fiction on all three layers, including emotional representation (3.36 vs.\ 4.04). That the deficit persists even against formulaic genre fiction, which is distinguished from canonical literature by critical reception rather than by literary training data volume, points toward something more fundamental than data insufficiency.

\subsection{Cross-Model Validation}

To assess whether reliability reflects the multi-round architecture rather than GPT-5-mini specifically, we replicated evaluations on a 30-story subset using Claude Sonnet via OpenRouter (59 of 60 story-mode pairs completed; one story excluded due to API timeout). Pearson correlations between GPT-5-mini and Claude scores range from $r$=0.636--0.812 (overall $r$=0.775, Spearman $\rho$=0.751--0.799). Claude scores systematically lower by $\approx$0.5 points (MAD 0.61--0.71), reflecting inter-model calibration differences rather than ranking disagreement; ranking agreement, not absolute scale alignment, is the relevant metric for cross-model validity. Full per-layer results are in Appendix~\ref{sec:appendix_crossmodel}.

\subsection{Dimensional Independence (RQ4)}

Cross-layer correlations (Table~\ref{tab:correlation}) are moderate and positive ($r$=0.649--0.683, all $p<0.001$), falling just below the conventional redundancy threshold of $r>0.7$; the margin is modest but consistent across all three pairs, supporting the claim that layers capture non-redundant aspects of literary quality. The distinct overall means reinforce this: L5 Emotional (3.96) $>$ L4 Cultural (3.64) $>$ L6 Existential (3.59), reflecting that emotional patterns are more broadly distributed across authorship types, while cultural and philosophical dimensions differentiate more sharply by genre.

\begin{table}[t]
\centering
\caption{Cross-layer correlation analysis ($N$=200).}
\label{tab:correlation}
\small
\resizebox{\columnwidth}{!}{%
\begin{tabular}{lccc}
\toprule
\textbf{Layer Pair} & \textbf{Pearson $r$} & \textbf{Spearman $\rho$} & \textbf{$p$} \\
\midrule
L4 $\times$ L5 & 0.649 & 0.514 & $<$0.001 \\
L4 $\times$ L6 & 0.683 & 0.598 & $<$0.001 \\
L5 $\times$ L6 & 0.680 & 0.621 & $<$0.001 \\
\bottomrule
\end{tabular}%
}
\end{table}

\section{Discussion}
\label{sec:discussion}

\subsection{The Capability Boundary of LLM-Generated Narratives}

The differential effect sizes reported in Section~\ref{sec:results} (L4 $d$=2.68 and L6 $d$=2.40 vs.\ L5 $d$=1.68) reveal a genuine asymmetry in what current generative models can and cannot reproduce. We interpret this asymmetry through the distinction between \emph{pattern-reproducible} and \emph{stance-requiring} literary capacities. Emotional structures such as affective complexity, psychological interiority, and emotional granularity are densely represented in narrative training corpora. Characters experience recognizable emotions; established literary techniques convey interiority; emotional arcs follow learnable patterns. LLMs trained on large fiction corpora can reproduce these surface structures with moderate fidelity (LLM mean 3.36 vs.\ canonical 4.15). Cultural power critique and existential-philosophical inquiry, by contrast, require more than pattern matching: a critical stance toward the social structures being represented, original engagement with questions of finitude and meaning, and the capacity to position narrative voice within contested ideological terrain. If the pattern-reproducible/stance-requiring distinction is correct, we would expect emotional dimensions---where surface patterns in training data are dense---to show smaller gaps than cultural and philosophical ones, where pattern matching provides less purchase. The approximately doubled effect sizes for L4 and L6 are consistent with this prediction, suggesting qualitatively different challenges rather than merely harder instances of the same task. While alternative explanations exist (for instance, that emotional dimensions exhibit lower variance across text types, compressing effect sizes mechanically), the pulp fiction comparison provides a more direct test.

The pulp fiction results sharpen this interpretation in a way the canonical-vs.-LLM comparison alone cannot. Pulp fiction authors consume the same literary training signal as LLMs---genre fiction is abundant in any large corpus---so if the LLM deficit were primarily a matter of data volume or literary exposure, pulp fiction would be the category where gaps should be smallest. Instead, human commercial authors, writing for mass markets under genre constraints, nonetheless achieve emotional scores approaching canonical literature (4.04 vs.\ 4.15, gap of just 2.7\%), while LLM-generated stories fall \emph{below} pulp fiction on all three layers, including emotional representation (3.36 vs.\ 4.04). The cultural and philosophical gaps are even larger (+1.28 and +1.09 points against pulp). This pattern cannot be explained by data insufficiency alone. Pulp fiction authors, however formulaic, write from particular social positions and make choices about how to represent cultural experience; they take stances, even conventional ones. The LLM deficit in cultural and philosophical dimensions is consistent with an absence of such positioning, though whether this reflects a fundamental architectural limitation or a training signal deficit remains an open question.

The cross-layer correlation structure corroborates this account. The strongest association links cultural and existential dimensions ($r$=0.683), consistent with both requiring a critical stance toward represented material. The weakest links cultural and emotional dimensions ($r$=0.649), consistent with emotional competence being more broadly distributed across authorship types. Collapsing these into an aggregate quality score would obscure the very differentiation that reveals the capability boundary.

\subsection{Why Reliable Evaluation of These Dimensions Matters for NLG}

The capability boundary identified above would be invisible to existing NLG evaluation methods (see Section~\ref{sec:related}). The pattern-reproducible dimensions where LLMs perform best (fluency, surface coherence, emotional mimicry) are precisely those that existing metrics already capture reasonably well. The stance-requiring dimensions where LLMs perform worst (cultural critique, $d$=2.68; philosophical depth, $d$=2.40) are the dimensions that existing metrics cannot see at all. The consequence is a systematic blind spot in the feedback available to NLG systems: improvement signals are rich where improvement is least needed, and absent where the gaps are largest.

SAGE addresses this blind spot by providing theory-grounded, reliable evaluation signal for cultural, emotional-psychological, and existential-philosophical dimensions. The reliability demonstrated here (98.8\% convergence, $>$94\% inter-rater agreement, near-perfect mode invariance) is a prerequisite for using evaluation as a training or feedback signal. An evaluator that produces unstable or inconsistent scores on these dimensions cannot guide improvement; an evaluator that achieves measurement-grade reliability can. Prior LLM-as-judge work demonstrates viability in conversational and instructional settings \citep{zheng2023judging,li2024llms}; our results extend this to interpretive literary dimensions considerably more complex, showing that iterative self-reflection, layer-specific bias mitigation, and independent cross-validation are sufficient to achieve comparable reliability on tasks that single-pass evaluation cannot handle. Cross-model validation with Claude Sonnet (overall $r$=0.775, Section~\ref{sec:results}) confirms this reliability is a property of the architecture, not the specific model.

\subsection{Scope and Limitations}

\paragraph{Corpus and generalizability.} The evaluation corpus comprises 100 English-language short stories; generalizability to other languages, narrative lengths, and cultural traditions requires separate validation. The framework validates only the interpretive layers (L4--L6); full integration with the rule-based L1--L3 metrics is left for future work. Notably, the canonical vs.\ pulp fiction distinction is not statistically significant on any layer ($p>0.05$), indicating that SAGE's discriminative sensitivity is stronger at the human-vs.-LLM boundary than within human authorship tiers; finer-grained discrimination may require larger corpora or additional layers.

\paragraph{External validity.} The inter-rater agreement exceeding 94\% reflects consistency between two LLM-based evaluators rather than agreement with professional literary critics. As a proxy for external validity, we test directional consistency against scholarly critical consensus: for 15 story pairs with well-established critical reputations, SAGE rankings match scholarly consensus in 13 of 15 cases (87\%), with perfect accuracy on L5 and L6 (5/5 each, $p$=0.031 binomial test). The two discordant cases in L4 are interpretively interesting rather than merely anomalous: Lu Xun (canonical Chinese modernist fiction) scores below Carver (working-class American realism) and Chambers (fin-de-si\`{e}cle Gothic fiction) on cultural representation. Carver's precise social observation and Chambers's cultural anxiety register as culturally dense on the dimensions SAGE measures, consistent with scholarly readings that identify specificity in these works that canonical criticism has underemphasized. Full validation against professional literary critics remains essential.

\section{Conclusion}
\label{sec:conclusion}

We introduced SAGE, a hierarchical framework for evaluating interpretive literary quality in narratives through structured decomposition into theory-grounded dimensions assessed via multi-round iterative LLM evaluation with independent cross-validation. Validated across 600 evaluations, the framework achieves measurement-grade reliability (convergence, inter-rater agreement, and mode invariance all meeting or exceeding thresholds), confirmed as a property of the multi-round architecture rather than any single model.

The central empirical finding is a systematic capability boundary: LLM-generated narratives approach human competence on emotional-psychological representation, but exhibit approximately double the gap on cultural critique and philosophical depth. The fact that LLM-generated stories score below even commercial genre fiction on all three layers points toward an absence of genuine epistemic positioning rather than insufficient literary training. We characterize this as a boundary between pattern-reproducible literary capacities, which LLMs can learn from surface regularities in training data, and stance-requiring ones, which demand cultural positioning and philosophical engagement that pattern matching alone cannot provide.

Several limitations constrain generalizability. The corpus is restricted to 100 English-language short stories; cross-lingual and cross-cultural validation is needed. Inter-rater agreement reflects LLM-to-LLM consistency rather than agreement with human literary experts; while directional consistency with scholarly critical consensus reaches 87\% (13/15 story pairs, $p$=0.031), external validation against professional critics remains essential. Future work should examine whether evaluation feedback on stance-requiring dimensions can be integrated into NLG training pipelines, and whether the capability boundary identified here proves stable or shifts as generative architectures advance.

\section*{Acknowledgements}

\bibliography{references}

\section*{Ethical Considerations}

\paragraph{Use of literary texts.}
The evaluation corpus comprises 100 short stories across three categories. Canonical and pulp fiction works published before 1928 are in the public domain and freely available from Project Gutenberg, Wikisource, and the Internet Archive. Works published between 1928 and 1990 that remain under copyright are used under the Fair Use doctrine (17 U.S.C.\ \S 107) for non-commercial academic research. This use is transformative (literary texts are converted into numerical evaluation scores across interpretive dimensions, not reproduced or redistributed) and consistent with Fair Use precedent in computational literary studies \citep[Authors Guild v. HathiTrust, 2014;][Authors Guild v. Google, 2015]{}. No copyrighted texts are included in supplementary materials or the public repository; only evaluation scores and aggregate statistics are released.

\paragraph{LLM-generated stories.}
The 20 LLM-generated stories are drawn from the publicly available \texttt{lars76/story-evaluation-llm} dataset \citep{huggingface-story-eval} under its stated license. No personally identifiable information is present in this dataset.

\paragraph{LLM-based evaluation.}
Automated evaluation using LLMs carries the risk of systematic bias, including cultural projection, Western-centric norms, and hallucinated textual evidence. SAGE's multi-round architecture includes explicit bias mitigation at Round~2 (projection bias and hallucination checks) and Round~3 (layer boundary enforcement). Residual bias cannot be fully eliminated; results should be interpreted as evidence from a structured automated evaluation system rather than as ground truth literary judgments.

\paragraph{Data availability.}
Evaluation scores for all 600 evaluations, prompt templates, analysis scripts, and the story catalog (with metadata but not copyrighted text) are publicly available at \url{https://anonymous.4open.science/r/sageFramework-journal-EDD6/}.

\appendix
\section{Exemplar Matrices}
\label{sec:appendix_exemplars}

Tables~\ref{tab:l4_exemplars}--\ref{tab:l6_exemplars} present the full exemplar matrices for Layers 4, 5, and 6, demonstrating construct validity through canonical literary contrasts. Each table illustrates that all four dimensions within the layer are non-redundant: works can achieve high scores on one dimension while scoring low on another, confirming that the dimensions capture genuinely independent aspects of literary quality.

\begin{table}[ht]
\centering
\caption{L4 exemplar matrix. H\,=\,High, M\,=\,Medium, L\,=\,Low.}
\label{tab:l4_exemplars}
\small
\setlength{\tabcolsep}{5pt}
\resizebox{\columnwidth}{!}{%
\begin{tabular}{p{3.6cm}lcccc}
\toprule
\textbf{Work} & \textbf{Form} & \textbf{IPD} & \textbf{CVP} & \textbf{CSP} & \textbf{CPC} \\
\midrule
\emph{Things Fall Apart} (Achebe) & Novel & H & H & H & H \\
\emph{Heart of Darkness} (Conrad) & Novel & H & L & M & M \\
\emph{The Great Gatsby} (Fitzgerald) & Novel & L & L & H & L \\
\emph{Animal Farm} (Orwell) & Novel & H & L & L & L \\
``Everyday Use'' (Walker) & Short & H & H & H & H \\
``Lady with the Dog'' (Chekhov) & Short & L & L & H & L \\
``The Dead'' (Joyce) & Short & M & H & H & H \\
``Tl\"{o}n, Uqbar...'' (Borges) & Short & L & L & M & H \\
\bottomrule
\end{tabular}%
}
\end{table}

IPD and CSP vary independently: Orwell's \emph{Animal Farm} maximizes power visibility while stripping cultural specificity to serve allegory; Fitzgerald's \emph{The Great Gatsby} renders Jazz Age New York with ethnographic precision while treating aspiration rather than power asymmetry as its central subject. CVP and IPD dissociate: Conrad and Achebe share comparable IPD while occupying opposite CVP positions, since Conrad renders Africa through a European observer's consciousness while Achebe renders the same colonial encounter from within Igbo life. CPC can reach high values with low IPD, as in Borges's ``Tl\"{o}n,'' which stages incommensurable epistemic systems without organizing them through power hierarchy.

\begin{table}[ht]
\centering
\caption{L5 exemplar matrix. H\,=\,High, M\,=\,Medium, L\,=\,Low. $^\dagger$EG operates at narrative apparatus level, not character self-report.}
\label{tab:l5_exemplars}
\small
\setlength{\tabcolsep}{5pt}
\resizebox{\columnwidth}{!}{%
\begin{tabular}{p{3.8cm}lcccc}
\toprule
\textbf{Work} & \textbf{Form} & \textbf{AC} & \textbf{PI} & \textbf{EG} & \textbf{ENC} \\
\midrule
``Hills Like White Elephants'' (Hemingway) & Short & H & L & L & H \\
``The Metamorphosis'' (Kafka) & Short & M & L & L & H \\
``Runaway'' (Munro) & Short & H & H & H & H \\
``Miss Brill'' (Mansfield) & Short & M & H & H$^\dagger$ & H \\
``Cathedral'' (Carver) & Short & L$\to$H & L & L & H \\
\emph{Mrs.\ Dalloway} (Woolf) & Novel & H & H & H & H \\
\emph{In Search of Lost Time} (Proust) & Novel & H & H & M & H \\
\emph{Neapolitan Novels} (Ferrante) & Novel & H & H & H & H \\
\bottomrule
\end{tabular}%
}
\end{table}

AC and PI dissociate: Hemingway's ``Hills Like White Elephants'' stages extreme emotional complexity while providing no access to either character's inner states (AC high, PI near zero); Woolf's \emph{Mrs.\ Dalloway} achieves both simultaneously through stream-of-consciousness technique. EG and ENC dissociate in Carver's ``Cathedral'': the narrator employs deliberately flat emotional vocabulary throughout yet achieves high ENC because the narrator's transformation is fully motivated by the marriage, the defensiveness, and Robert's patience that the narrative has established.

\begin{table}[ht]
\centering
\caption{L6 exemplar matrix. H\,=\,High, M\,=\,Medium, L\,=\,Low. $^*$LP structurally inaccessible (clones possess no future). $^\dagger$ME collapses at death but constitutes the story's subject throughout. $^\ddagger$MR suspended: narrative refuses moral adjudication.}
\label{tab:l6_exemplars}
\small
\setlength{\tabcolsep}{4pt}
\resizebox{\columnwidth}{!}{%
\begin{tabular}{p{3.8cm}lcccc}
\toprule
\textbf{Work} & \textbf{Form} & \textbf{LP} & \textbf{MR} & \textbf{HC} & \textbf{ME} \\
\midrule
\emph{The Stranger} (Camus) & Novel & H & L & H & H \\
\emph{Crime and Punishment} (Dostoevsky) & Novel & H & H & H & H \\
\emph{Germinal} (Zola) & Novel & L & M & H & L \\
\emph{Middlemarch} (Eliot) & Novel & L & H & H & M \\
\emph{Never Let Me Go} (Ishiguro) & Novel & L$^*$ & M & H & H \\
``The Guest'' (Camus) & Short & H & Susp.$^\ddagger$ & H & H \\
``A Clean, Well-Lighted Place'' (Hemingway) & Short & H & L & H & H \\
``A Hunger Artist'' (Kafka) & Short & H & L & H & H$^\dagger$ \\
``The Dead'' (Joyce) & Short & L & L-M & H & H \\
\bottomrule
\end{tabular}%
}
\end{table}

LP and MR dissociate in both directions: Camus's \emph{The Stranger} achieves high LP through Meursault's coherent absurdist life-position while MR is near zero, since the novel's power depends on refusing moral evaluation. Dostoevsky's \emph{Crime and Punishment} shares high LP while MR constitutes its entire subject. HC and ME dissociate: Zola's naturalism and Hemingway's ``A Clean, Well-Lighted Place'' share maximal HC but diverge on ME because Zola's deterministic framework forecloses meaning from within while Hemingway's old waiter constructs meaning through maintenance of order against the nada.

\section{Evaluation Protocol and Prompt Templates}
\label{sec:appendix_prompts}

\subsection*{Five-Round Iterative Protocol}

\begin{table}[ht]
\centering
\caption{Five-round iterative evaluation protocol (all layers).}
\label{tab:eval_protocol}
\small
\resizebox{\columnwidth}{!}{%
\begin{tabular}{clp{5.5cm}}
\toprule
\textbf{R} & \textbf{Focus} & \textbf{Task} \\
\midrule
1 & Initial & Extract layer-relevant content; score all 4 dimensions (1.0--5.0) with confidence and textual evidence \\
2 & Bias check & L4: hallucination check. L5: projection bias (emotional norms). L6: Western framework imposition \\
3 & Boundary & Verify no cross-layer contamination \\
4 & Evidence & Adjust \emph{confidence} (not scores) if evidence is sparse or ambiguous \\
5 & Final & Confirm or adjust scores; convergence: $|\Delta|<0.3$ from R4 to R5 \\
\bottomrule
\end{tabular}%
}
\end{table}

\subsection*{Scoring Rubric}

\begin{table}[ht]
\centering
\caption{Scoring rubric applied uniformly across all layers.}
\label{tab:scoring_rubric}
\small
\resizebox{\columnwidth}{!}{%
\begin{tabular}{cp{6.2cm}}
\toprule
\textbf{Score} & \textbf{Criteria} \\
\midrule
4.5--5.0 & Exceptional: rich, multi-layered representation; extensive evidence; demonstrable mastery \\
4.0--4.5 & Strong: substantial representation; clear evidence; depth and nuance \\
3.0--3.5 & Moderate: some representation; adequate but not extensive evidence \\
2.0--2.5 & Weak: minimal representation; surface-level; sparse evidence \\
1.0--1.5 & Absent/Negligible: dimension effectively absent; no meaningful evidence \\
\bottomrule
\end{tabular}%
}
\end{table}

\subsection*{Prompt Design Summary}

Full prompt templates and JSON schemas for all three layers and both modes are available at \url{https://anonymous.4open.science/r/sageFramework-journal-EDD6/prompts/prompt\_templates.md}. Table~\ref{tab:prompt_design} summarizes the key design decisions instantiated in each prompt component.

\begin{table}[ht]
\centering
\caption{Prompt design: components and layer-specific instantiations.}
\label{tab:prompt_design}
\small
\resizebox{\columnwidth}{!}{%
\begin{tabular}{p{2.2cm}p{5.8cm}}
\toprule
\textbf{Component} & \textbf{Design} \\
\midrule
System role & Expert analyst for the target layer; dimension definitions with theoretical citations \\
\midrule
Core principles & (1) Evidence Sovereignty: all scores cite textual evidence. (2) Layer Boundary: evaluate only target-layer content. (3) Projection Awareness: no norm imposition. (4) Dimension Independence: score each dimension separately \\
\midrule
Mode instruction & Content-limit: text only, no prior knowledge. Title-limit: scholarly consensus from title/author only \\
\midrule
R1 output schema & JSON: extracted content, per-dimension score (1.0--5.0) / confidence (1--5) / reasoning / evidence strength, overall score \\
\midrule
R2 bias check & \textit{L4}: hallucination check (fabricated cultural/historical claims). \textit{L5}: projection bias (Western emotional norms). \textit{L6}: framework imposition (Sartre/Camus onto non-Western texts; overlooking Buddhist/Confucian/Daoist depth) \\
\midrule
R3 boundary & Verify no cross-layer contamination (e.g., L5 must not score cultural power or philosophical claims) \\
\midrule
R4 evidence & Re-examine evidence quality; adjust confidence only (not scores) if evidence is sparse \\
\midrule
R5 final & Consolidate scores; flag convergence ($|\Delta| < 0.3$ from R4 to R5) \\
\midrule
Validator role & Independent scoring + bias/hallucination detection + agreement analysis; no access to iterative evaluator outputs \\
\bottomrule
\end{tabular}%
}
\end{table}


\section{Convergence Trajectories}
\label{sec:appendix_convergence}

\begin{figure}[ht]
\centering
\includegraphics[width=\columnwidth]{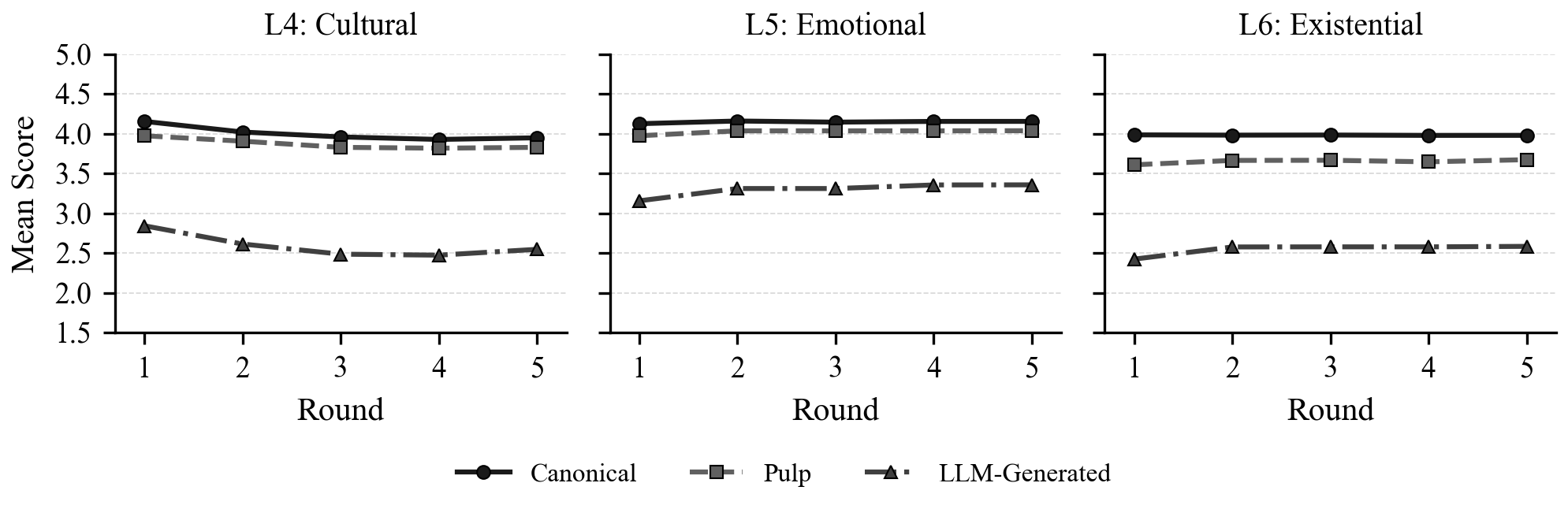}
\caption{Score convergence trajectories across five iterative rounds for each layer (L4--L6) and genre category. Scores stabilize by Rounds~3--4; canonical and pulp fiction consistently score higher than LLM-generated stories.}
\label{fig:convergence}
\end{figure}

\begin{figure}[ht]
\centering
\includegraphics[width=\columnwidth]{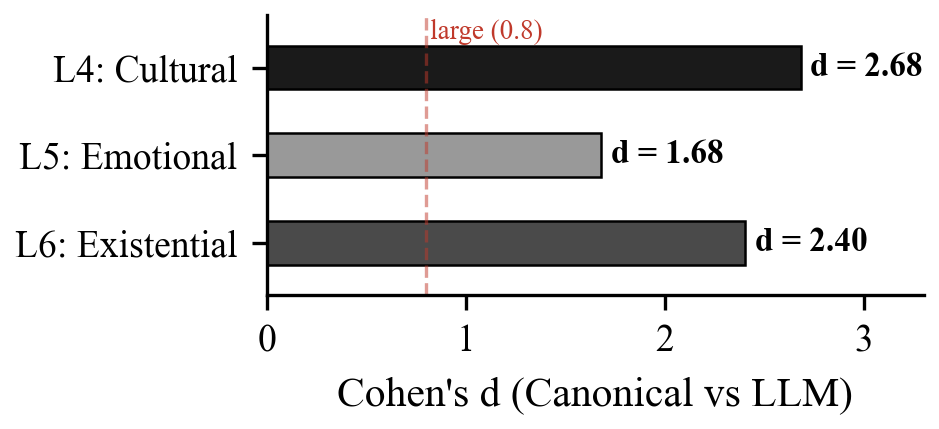}
\caption{Effect size (Cohen's $d$) for Canonical vs.\ LLM-generated stories across three layers. All values exceed the large-effect threshold ($d=0.8$, dashed line).}
\label{fig:effects}
\end{figure}

Figure~\ref{fig:convergence} shows score trajectories across the five iterative rounds for each layer and genre category; Figure~\ref{fig:effects} shows the corresponding effect sizes for the canonical vs.\ LLM-generated comparison.

\section{Cross-Model Validation Results}
\label{sec:appendix_crossmodel}

\begin{table}[ht]
\centering
\caption{Cross-model consistency: GPT-5-mini vs.\ Claude Sonnet on 30-story subset (59 pairs per layer).}
\label{tab:crossmodel}
\small
\resizebox{\columnwidth}{!}{%
\begin{tabular}{lcccc}
\toprule
\textbf{Layer} & \textbf{Pearson $r$} & \textbf{Spearman $\rho$} & \textbf{MAD} & \textbf{GPT $\mu$ / Claude $\mu$} \\
\midrule
L4 Cultural    & 0.812 & 0.799 & 0.705 & 3.46 / 2.87 \\
L5 Emotional   & 0.636 & 0.751 & 0.622 & 3.83 / 3.39 \\
L6 Existential & 0.797 & 0.778 & 0.614 & 3.44 / 2.97 \\
\midrule
\textbf{Overall} & \textbf{0.775} & --- & \textbf{0.647} & --- \\
\bottomrule
\end{tabular}%
}
\end{table}

\end{document}